\documentclass{article} %
\usepackage{iclr2027_conference,times}

\usepackage{amsmath,amsfonts,bm}

\def\eqref#1{equation~\ref{#1}}

\def\1{\bm{1}}

\DeclareMathAlphabet{\mathsfit}{\encodingdefault}{\sfdefault}{m}{sl}
\SetMathAlphabet{\mathsfit}{bold}{\encodingdefault}{\sfdefault}{bx}{n}

\usepackage{hyperref}
\usepackage{url}
\usepackage{amssymb}
\usepackage{amsthm}
\usepackage[utf8]{inputenc} %
\usepackage[T1]{fontenc}    %
\usepackage{courier}
\usepackage{url}            %
\usepackage{booktabs}       %
\usepackage{amsfonts}       %
\usepackage{nicefrac}       %
\usepackage{microtype}      %
\usepackage{multirow}
\usepackage{amsmath}
\usepackage{graphicx}
\usepackage{enumitem}
\usepackage{wrapfig}
\usepackage{xcolor}
\usepackage{float}
\newtheorem{definition}{Definition}
\newsavebox{\FullLUTBox}
\newsavebox{\DiagnosisBox}
\newlength{\DiagnosisRowExtra}

\title{Bits Under ZK-LLM: \\Evaluating Zero-Knowledge-Friendly Quantization for Verifiable Private LLM Inference}

\author{Taeung Yoon, Yupeng Zhang, Xiaojing Liao \\
Siebel School of Computing and Data Science\\
University of Illinois Urbana-Champaign\\
\texttt{\{tuyoon2,zhangyp,xjliao\}@illinois.edu} \\
}

\iclrfinalcopy %
\begin{document}

\maketitle

\begin{abstract}

Zero-knowledge (ZK) proofs are emerging as a promising approach for enabling private, verifiable LLM governance and auditing, where regulators, users, and auditors need to verify claims about training-data usage or LLM inference-time behavior, while model providers must protect proprietary model parameters.
However, despite the growing interest in
ZK-LLMs, the understanding of ZK-friendly quantization remains limited. 
This gap matters because in the ZK setting, quantization directly shapes the arithmetic structure, constraint complexity, and proving cost of ZK inference. ZK protocols operate over finite fields and incur costs that depend heavily on the number and type of arithmetic operations, nonlinearities, and lookup constraints.
Understanding ZK-friendly quantization is therefore essential for making ZK-LLMs practical.
In this work, we present the first systematic study of ZK-friendly quantization for LLMs. 
We first formalize the definition of ZK-friendly quantization, capturing the properties required for ZK proof generation. 
We then evaluate nine language models, including Qwen2.5-14B and the mixture-of-experts model Qwen3-30B-A3B, across a broad design space of weight, activation, and nonlinear lookup table precision.  
Our results show that activation precision is substantially more sensitive than weight precision, while nonlinear lookup approximations can become the dominant source of utility degradation.
Also, we identify RMSNorm inverse-square-root lookups as a recurring bottleneck in several large models and recover near-baseline utility by selectively increasing precision only at the bottleneck.
Finally, we show that reducing bit-width or lookup-table size does not necessarily yield proportional end-to-end proving savings, showing that conventional low-bit quantization heuristics do not directly translate to ZK proving efficiency and motivating operator-aware precision selection.

\end{abstract}

\section{Introduction}

Large language models (LLMs) are increasingly deployed through opaque cloud services, yet users, auditors, and regulators often lack practical mechanisms to verify what computation was performed, which model was used, or whether specific governance-relevant claims hold. 
Zero-knowledge (ZK) proofs offer a promising path toward verifiable private LLM inference~\citep{sun2024zkllm, qu2025zkgpt, xie2025zkpytorch, chen2024zkml, liao2025verilora,wang2026zkagent,gailly2026deepprove}: a provider can prove that an inference-time computation or auditing statistic was correctly computed without revealing proprietary model weights or sensitive operational details. 
However, proving LLM inference directly is costly because ZK proof systems operate over finite fields, while modern transformers rely on floating-point arithmetic, nonlinear functions, normalization, and softmax operations that are expensive to arithmetize. 
Compared with conventional machine learning (ML) quantization, ZK-friendly quantization is therefore not only a compression technique; it determines the arithmetic structure, constraint complexity, and proving cost of the resulting ZK-LLM system.

Despite rapid progress in ZK-ML and LLM quantization, there is still no formal definition of  ZK-friendly quantization or a systematic understanding of which quantization schemes are actually suitable for ZK-LLM. 
Prior ZK-LLM systems typically adopt a specific integer representation or lookup approximation as an implementation choice, often validating it on a limited model family or scale~\citep{sun2024zkllm, qu2025zkgpt}, while conventional LLM quantization benchmarks optimize for deployment efficiency rather than proof compatibility~\citep{ yang2024llmcbench, huang2024good}. 
This leaves open a fundamental research gap: we do not yet know how ZK-friendly quantization mechanisms affect ZK proof generation across models, tasks, and operator types. 
This gap matters because, in the ZK setting, quantization directly determines the arithmetic structure of the proved computation, including its constraint complexity, lookup requirements, and prover cost. In addition, these trade-offs directly affect governance applications that must preserve model behavior while remaining practical to prove. We demonstrate this in a ZK training-data membership case study based on Min-K\% inference (Appendix~\ref{sec:membership}).
Note that although this study focuses on ZK-LLMs, these considerations may also extend to other cryptographic approaches for private or verifiable inference, such as secure multiparty computation~\citep{yao1982protocols, hou2026ciphergpt} and fully homomorphic encryption~\citep{gentry2009fully}, as they often represent computations over finite fields or polynomial rings and therefore face similar challenges.
Understanding ZK-friendly quantization is essential for making proof systems practical.

\textbf{Contributions}. 
This study makes three major contributions toward understanding and designing ZK-friendly quantization for LLMs.
First, we formalize \emph{ZK-friendly quantization} as a joint utility-proof-efficiency problem, capturing the constraints imposed by bounded integer representations, fixed rescaling rules, static calibration, and lookup-based nonlinear computation. Building on this formulation, we develop a scalable methodology that enables systematic study across model scales and architectures.
Second, we show that the utility bottlenecks of ZK-friendly quantization differ substantially from those of traditional LLM quantization. Across nine language models, we find that model utility is markedly more sensitive to activation precision than weight precision, while fixed-precision nonlinear lookup table (LUT) can dominate utility degradation even when linear quantization remains accurate. Also importantly, we identify the inverse-square-root LUT in RMSNorm as a recurring bottleneck in several large models and show that selectively increasing its LUT precision can recover near-baseline utility without uniformly increasing precision across all nonlinearities.
Third, we characterize how these quantization choices translate into realized ZK proving cost. Using direct proof-system measurements, we find that reducing weight/activation bit-width or LUT size does not necessarily yield proportional ZK prover savings. These results motivate heterogeneous, operator-aware precision choices rather than uniformly minimizing bit-width, and expose the utility-proving-cost trade-offs that determine practical ZK-LLM inference.

\section{Background and Related Work}
\label{sec:background}

\paragraph{Zero-Knowledge Proofs.}

Zero-knowledge proofs (ZKPs) are cryptographic protocols that allow a prover to
convince a verifier that a computation was executed correctly without revealing
the secret witness. In ML, a model provider can use a ZKP to
prove that an inference or other properties, such as fairness, robustness, and explainability~\citep{yadav2024fairproof,zhang2025fairzk,yadav2026expproof}, are honestly computed on the secret model~\citep{feng2021zen, lee2024vcnn, chen2024zkml}.
ZKP protocols represent computations as arithmetic operations defined over a large prime
field $\mathbb{F}_{p}$. The natively supported operations include modular additions, $a+b \mod p$, and modular multiplications, $a \times b \mod p$ for $a,b\in\mathbb{F}_p$. 
Modern proof systems also support lookup arguments \citep[e.g.,][]{setty2024unlocking} that check whether an input-output pair belongs to a fixed table, denoted as $y
\leftarrow \mathsf{Lookup}(T,x)$. 
We use the term ``$B$-bit LUT'' to denote a table
with $2^B$ entries.
The proving cost depends on the table size and the total number of lookup operations.
Meanwhile, divisions are not native field operations. Also, representing floating-point arithmetic or continuous functions inside a proof requires extra constraints, such as range checks, bit decomposition, or lookup arguments. These overheads make ZK-LLM inference expensive. 

\paragraph{Post-Training Quantization.}

Post-training quantization (PTQ) is an approach to reduce the storage and memory cost of LLMs by lowering the bit-width of weights and/or activations~\citep{dettmers2023qlora, yang2024llmcbench, dettmers2022gpt3, xiao2023smoothquant, jacob2018quantization, frantar2022gptq}. 
In a basic form of quantization, a tensor $X$ is assigned a scale
$
    s = \frac{\max |X|}{2^{b-1}-1},
$
where $b$ is the target bit width. 
Each tensor value $x$ is then mapped to a signed $b$-bit integer by rounding $x/s$ and clipping to the representable range.
Static quantization computes the scale with calibration samples, whereas dynamic quantization computes it from runtime values~\citep{xie2025zkpytorch, xiao2023smoothquant}. 
A line of work~\citep{kurtic2025give, yang2024llmcbench, yao2023zeroquantv2exploringposttrainingquantization, huang2024good, gong2024llmcbenchmarkinglargelanguage} has explored how different quantization schemes affect performance and cost efficiency. LLMCBench~\citep{yang2024llmcbench} presents a benchmark for LLM compression algorithms, including seven modern quantization methods~\citep{ma2023llm, sun2023simple, frantar2023sparsegpt, frantar2022gptq, xiao2023smoothquant, lin2024awq, shao2023omniquant}, across multiple metrics such as compression performance. \citet{kurtic2025give} investigate the accuracy and cost trade-offs of specific quantization formats, such as W4A16, on both academic and real-world benchmarks, and show that W4A16-INT is one of the most cost-efficient settings. 
However, these benchmarks are designed for general ML environments, rather than the arithmetic constraints and proving costs of ZK protocols.
In ZK protocols, rescaling and nonlinear operations must also be encoded using field arithmetic, lookup arguments, or other proof-compatible constructions.

\paragraph{Quantization in ZK-ML.}
Early ZK-ML frameworks targeted relatively simple neural networks~\citep{chen2024zkml,liu2021zkcnn, lee2024vcnn, feng2021zen} and recent systems such as zkGPT~\citep{qu2025zkgpt} and zkLLM~\citep{sun2024zkllm} extend proof generation to LLM inference with protocol-level optimizations. In many existing ZK machine learning schemes, quantization is an essential building block, as otherwise the overhead of simulating floating-point arithmetic would be too high in ZK protocols. Many of the systems~\citep{liu2021zkcnn,feng2021zen, sun2024zkllm,qu2025zkgpt} use integer representations by adopting an affine quantization scheme~\citep{jacob2018quantization}. Rescaling is performed after multiplication to avoid overflow~\citep{qu2025zkgpt} in $\mathbb{F}_p$. 
To avoid expensive operations, they also use approximation techniques such as LUTs or polynomial approximations.
Specifically, zkGPT~\citep{qu2025zkgpt} uses high-precision quantization and integer-ratio rescaling for GPT-2 Small~\citep{radford2019language}, while zkLLM~\citep{sun2024zkllm} uses specialized attention arithmetization and lookup arguments for nonlinear operations. However, the quantization was only tested on the small model evaluated in each paper, and the techniques may not generalize to other models. Moreover, existing implementations of ZK machine learning are optimized for proof generation but not for fast plaintext inference. They are mostly written in C++ and Rust, and lack support for LLM acceleration. Therefore, it is inefficient to test different quantization methods on different models. Our paper addresses this gap by developing a framework to implement different ZK-friendly quantization schemes and benchmarking their performance. This framework is important to develop new ZK-ML schemes in the future for larger models and different architectures (e.g., MoE) that are currently not supported by existing ZK-ML systems.

\section{ZK-friendly Quantization}\label{sec:definition}
In this section, we first introduce the terminology and syntax of ZK-friendly
quantization in general. We then describe the properties that make such a scheme
useful in practice.

Let $\mathcal{M}$ be a large language model with real-valued weights and forward pass computations (e.g., floating-point numbers and arithmetic).
For an input prompt $x$, $\mathcal{M}(x) \to y$ denotes the process
of running the model and producing an output $y$, such as next-token logits or a
sequence of tokens. 
The main object of a ZK-friendly quantization scheme is a quantization procedure
\[
    \mathsf{Quantize}(\mathcal{M}, \mathbb{F}_p) \to \mathcal{M}_{Q},
\]
which takes a real-valued model $\mathcal{M}$ and outputs a quantized model
$\mathcal{M}_{Q}$ whose inference computation is represented over $\mathbb{F}_p$ of the ZK protocol.
Any numerical formats, scaling rules, approximation ranges, LUTs and public constants are treated as part of
the quantization scheme. 

Specifically, we consider two basic quantization algorithms:

\begin{itemize}[leftmargin=*]
    \item $\mathsf{LinQuant}(\mathsf{L}, \mathbb{F}_p) \to
    \widetilde{\mathsf{L}}$ takes a real-valued linear tensor operation
    $\mathsf{L}$, such as matrix multiplication, convolution and $Q,K,V$ projections, and replaces it with a quantized operation
    $\widetilde{\mathsf{L}}$.

    \item $\mathsf{NonlinQuant}(g, \mathbb{F}_p) \to \widetilde{g}$ takes a
    real-valued nonlinear operation $g$, such as softmax, GeLU and normalization with inverse square root, and replaces it with a
    quantized arithmetization $\widetilde{g}$.
\end{itemize}

With these operations, the
real-valued computation of $\mathcal{M}$ is replaced by a computation
represented over $\mathbb{F}_p$. Then the inference under the quantized model can
be represented as:
$
    \mathcal{M}_{Q}(x) \to y_{Q}.
$

\paragraph{ZK efficiency.} 
In principle, any function that is computable by a computer can be realized in a ZK protocol. However, the overhead of modeling the function could be high. Therefore, a useful quantization scheme should improve the efficiency of the ZK protocol on the quantized model, compared to the original model and computation. Efficiency measures include prover time, proof size and verifier time. 
All three measures depend on the size of the relation $\mathcal{R}$ to be proven in ZK, which we use to formalize this property. 

We define $\mathcal{R}=\{(x,y);\mathcal{M}|\mathcal{M}(x)=y\}$ as the original relation of LLM inference with real-valued model $\mathcal{M}$ and $\mathcal{R}_{Q,p}=\{(x,y_Q);\mathcal{M}_Q|\mathcal{M}_Q(x)=y_Q\}$ as the quantized relation with the quantized model and computations. For a ZK protocol $\Pi$, we use $|\mathcal{R}|_\Pi$ to denote the size of the relation under $\Pi$, which is usually measured by the number of arithmetic operations over $\mathbb{F}_p$ and lookup operations. 

\begin{definition}
A quantization scheme has $\Delta$-ZK efficiency with respect to a ZK protocol $\Pi$, where  $\Delta = \frac{|\mathcal{R}_{Q,p}|_\Pi}{  |\mathcal{R}|_\Pi}$.

\end{definition}

A quantization scheme is interesting only if $\Delta<1$, which is often significantly smaller than 1 in practice. The speedup is also not smooth, as we will demonstrate in \S~\ref{sec:end-to-end-proving}.

\paragraph{Fidelity.} Quantization should not substantially degrade the quality of the model; otherwise the ZK proof would be meaningless in practice.

\begin{definition}
Let $d$ be a discrepancy measure between the two outputs, and let $\mathcal{D}$ be an input distribution of the prompts or token sequences. The
fidelity error of $Q$ on $\mathcal{M}$ is defined as:
\[
    \delta_{d}(\mathcal{M}, Q; \mathcal{D})
    =
    \mathbb{E}_{x \sim \mathcal{D}}
    \left[
        d\big(\mathcal{M}(x), \mathcal{M}_{Q}(x)\big)
    \right].
\]

\end{definition} 

Different choices of $d$ capture different levels of fidelity. At the numerical
level, $d$ may measure layer-wise error or logit distance. At the distributional
level, it may measure KL divergence between next-token distributions or
agreement among top-ranked tokens. At the task level, it may measure the gap in
perplexity, accuracy, or another downstream metric. We measure three levels of fidelity in \S~\ref{sec:eval}.

\textbf{Static compilability.} Almost all ZK schemes cannot handle dynamic relations efficiently. Therefore, when modeled as an arithmetic circuit/constraint system, the relation of the quantized model inference should not depend on the input known at runtime. It does not mean that all intermediate values are known in advance. Activations, attention scores, and other hidden states still depend on
the input. Rather, it means that the rules for representing and checking these
values are fixed before the proof is generated. This distinction is important
because input-dependent choices, such as computing new quantization parameters inside the forward pass, can
substantially increase the cost of the ZK proof.

\begin{definition}
A quantization scheme is statically compilable if in the quantized relation $\mathcal{R}_{Q,p}=\{(x,y_Q);\mathcal{M}_Q|\mathcal{M}_Q(x)=y_Q\}$, all arithmetic operations, public constants and lookup tables are fixed and independent of the runtime input $x$. 
\end{definition}

With the definitions above, the goal of a ZK-friendly quantization scheme is to minimize $\Delta$ while ensuring $\delta_d(\mathcal{M}, Q; \mathcal{D})\le \varepsilon$ for an acceptable threshold of fidelity $\varepsilon$. The static compilability is implicitly captured by $\Delta$ as it directly affects the size of $|\mathcal{R}_{Q,p}|$. 

\textbf{Why ZK-friendly quantization?} ZK-friendly quantization is a central systems primitive for practical zkML: by mapping floating-point inference to bounded integer arithmetic and lookup-compatible nonlinearities, it can substantially reduce proof complexity and proving cost. Without ZK-friendly quantization, the overhead of simulating floating-point arithmetic could make the proving time more than 1200$\times$ slower, as we demonstrate in \S~\ref{sec:end-to-end-proving}. However, prior ZK-LLM systems largely tailor quantization to a specific model and proof stack, leaving little guidance on how precision and LUT design should transfer across architectures and scales. We address this gap by systematically investigating the utility-proving-cost trade-offs in ZK-friendly quantizations.

\section{Methods}
\label{sec:framework}

\paragraph{Research Questions.}
The central research question of our study is:
\emph{How should ZK-friendly quantization be designed to make LLM inference practical to prove while preserving the utility of the original model?}
We decompose this question into three sub-questions.
(1) How much utility degradation does ZK-friendly quantization introduce when enabling LUT nonlinearities?
(2) What are the root causes of utility degradation, and what kind of ZK-friendly quantization best preserves model utility?
(3) How should ZK-friendly quantization be designed to achieve a favorable trade-off between model utility and ZK proving cost?

\paragraph{ZK-aware Quantization Framework.}
We develop a ZK-aware quantization framework for a suite of Transformer models, including commonly studied model scales and architectures in recent ZK-LLM work~\citep{qu2025zkgpt, sun2024zkllm}, and MoE architectures that are widely deployed, but have not been studied in the ZK-LLM literature.
Grounded in the criteria defined in \S~\ref{sec:definition}: a ZK-friendly quantization must use bounded integer representations, fixed rescaling rules, static calibration, and fixed LUTs so that the resulting inference relation can be compiled before proof generation.
Hence, given a pretrained model and a quantization configuration, the framework is designed to instrument the Transformer forward pass at the operation level to emulate ZK-compatible computation.
More specifically, linear tensor operations are replaced by bounded-integer quantize-dequantize modules, while selected nonlinear functions are replaced by fixed LUT modules.%

$\bullet$ \textit{ZK-aware quantization design space}.
Our design space includes five quantization axes: weight and activation bit-width, scale type, symmetry, activation calibration, and LUT precision. 
We write W$n$A$m$ for $n$-bit weight and $m$-bit activation quantization. 
For scale type, the framework supports standard floating-point scales, power-of-two scales, $c_1/c_2$ integer-ratio rescaling, and ZK-dedicated signed symmetric scales~\citep{gailly2026deepprove}.
It also supports symmetric and asymmetric quantization, as well as static-max, static-percentile, and dynamic activation calibration (see Appendix~\ref{sec:design_ablation}). 

Nonlinear operators introduce an additional design dimension because they are realized through LUTs rather than ordinary arithmetic. Therefore, the framework supports operator-specific LUT construction and substitution, allowing the input-address precision $B$ to be configured independently for each nonlinear operator type.
Throughout the experiments, we use the notation B16, for example, to denote a LUT with $2^{16}$ entries.
This is important because different nonlinearities exhibit distinct input ranges and approximation sensitivities, and therefore need not admit the same LUT configuration. 
This operator-level construction enables us to systematically explore heterogeneous LUT designs and identify which nonlinear approximations dominate the utility-proving-cost trade-off (see Appendix~\ref{sec:detailed_implementation}). %

$\bullet$ \textit{Operation coverage for ZK-LLMs}.
We instrument the Transformer forward pass at the level of the operations that affect ZK-compatible execution. Learned parameters, including attention and MLP projections, the language-modeling head, embeddings, and normalization affine parameters, are quantized according to the selected configuration. 
Intermediate activations are subjected to quantize-dequantize (QDQ) simulation at designated boundaries, including projection, normalization, activation-function, and residual-addition outputs. Specifically, QDQ maps each tensor to the bounded integer domain implied by the target precision and scaling policy and then dequantizes it for continued execution with standard PyTorch operators. For MoE architectures, we additionally capture expert and router projections, including functional linear calls that are not exposed as \texttt{nn.Linear} modules. Note that for QDQ to faithfully evaluate the utility of a ZK-compatible configuration, it must reproduce the numerical effect of the integer computation encoded by the prover. We establish this correspondence algebraically and validate it empirically (see details in  Appendix~\ref{sec:qdq_simulation}).

We handle nonlinear computation separately: GeLU/SiLU, the inverse square root in LayerNorm/RMSNorm, and the exponential and reciprocal components of softmax are replaced by precomputed LUTs under the chosen operator-specific precision. Operations that are not directly approximated by the quantization configuration, such as attention-score computation, context aggregation, RoPE, masking, reductions, and top-(k) selection, remain in floating-point arithmetic in the utility evaluator. This separation makes the approximation boundary explicit: it isolates the effects of the ZK-relevant numerical transformations while leaving unrelated computation unchanged.

\vspace{-5pt}
\paragraph{Scalable Estimation of ZK Proving Cost.}
Directly evaluating the ZK cost of every quantization configuration is impractical: each candidate must be realized as an integer computation graph and executed by a prover, making exhaustive exploration prohibitively expensive for large models. We address this challenge with a two-stage methodology that decouples large-scale quantization exploration from proof-system execution.
We first use the aforementioned QDQ simulation to instantiate arbitrary weight, activation, and nonlinear-LUT precisions with standard PyTorch operators and GPU acceleration.
This allows us to explore configurations, including unconventional precisions such as INT12, without implementing a dedicated integer kernel or ZK circuit for each candidate. Then, we map selected configurations to actual proof-system execution using DeepProve~\citep{gailly2026deepprove}. We extend DeepProve to independently control weight, activation, and nonlinear-LUT precision and obtain direct measurements of prover time, verifier time, and proof size. This two-stage methodology makes it possible to study the utility-proving-cost design space at model scales and quantization granularities that would be impractical to explore through proof generation alone.

\paragraph{Implementation and experiment setup.} We implement our approach in PyTorch and Hugging Face and evaluate nine language models spanning different scales and architectures, including a Mixture-of-Experts model. Models are loaded in BF16 and modified according to a specified quantization configuration. We measure language-modeling quality with perplexity on WikiText-2~\citep{merity2016pointer} and C4~\citep{raffel2020exploring}, downstream accuracy on ARC-Easy and MMLU using the \texttt{lm-evaluation-harness} multiple-choice protocol~\citep{gao2021framework}, and token-distribution fidelity to the BF16 model using KL divergence, Top-5 overlap, and bottom-20\% overlap (see Appendix~\ref{sec:behavioral_fidelity}).
For proof evaluation, we report prover time, verifier time, and proof size.
Weights are quantized post hoc, while activation scales are obtained from calibration data and then fixed for evaluation. Unless stated otherwise, we use signed symmetric per-tensor quantization with standard scales and static-max activation calibration over eight fixed WikiText-2 training windows. Auxiliary domains, including embeddings, positional representations, and residual paths, are kept at 24 bits while weight and activation precision are varied. QDQ modules are inserted at the activation boundaries, and nonlinear operators are replaced by calibrated LUTs at the selected LUT precision. Further implementation details and QDQ validation are provided in Appendices~\ref{sec:qdq_simulation} and~\ref{sec:lookup}.

\section{Results}
\label{sec:eval}

\subsection{Broad Utility Study}
\label{subsec:broad-utility}
\paragraph{R1. Activations are more sensitive to low precision than weights} 
We first isolate linear quantization, which accounts for most bounded-integer arithmetic in the ZK computation.
For activation sensitivity, we fix weight bit precision at 16 bits, and scale the bit precision of activations from 8 to 16, incrementing by 2. For weight sensitivity, we fix activations at 16 bits, and increase the bit precision of weights from 8 to 16 by 2.

\begin{figure}[htbp]
  \centering
  \includegraphics[width=.8\textwidth]{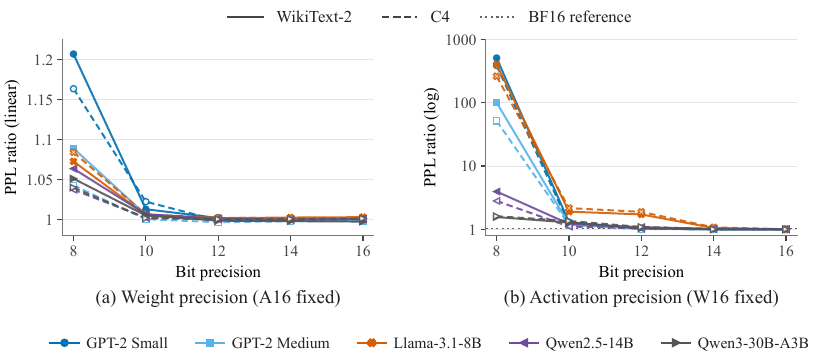}
  \vspace{-15pt}
  \caption{Perplexity ratio under linear quantization to BF16 baseline.}
  \label{fig:linear-quant}
\end{figure}

Figure~\ref{fig:linear-quant} shows whether weight or activation precision limits utility. In Figure~\ref{fig:linear-quant} (a), even at the lowest precision of 8 bits, the PPL ratio compared to the BF16 baseline remained below about 1.2. 
Figure~\ref{fig:linear-quant} (b) shows that most models exhibited PPL ratios above 10 at 8 bits. Most models required at least 12 or 14 bits for the ratio to approach 1. 
This sensitivity does not vary monotonically with model size, and the same trend holds across WikiText-2 and C4.
We provide the complete results in Appendix~\ref{app:additional-utility}.

These results show that activation quantization has greater impact on model utility than weight quantization. 
Since LUT approximations can only introduce additional error, 8-bit activations already leave too little utility margin for the full ZK-friendly pipeline.
Depending on the model, activation precisions of at least 12 or 14 bits provide a practical starting point for preserving utility.

\paragraph{R2. A uniform B16 LUT configuration does not reliably preserve utility}
After identifying the linear quantization regimes, we evaluate the full ZK-friendly pipeline by replacing nonlinear transformer operations with fixed LUTs. 
This experiment tests whether utility loss comes mainly from linear quantization, nonlinear approximation, or their interaction.

$\bullet$ \textit{Full-pipeline evaluation}.
Table~\ref{tab:lut-results} (a) reports perplexity ratios for the full pipeline.
GeLU or SiLU activations, normalization inverse square roots, softmax exponentials and reciprocals, including the additional routing-related functions in Qwen3 are replaced by a $B$-bit LUT over a calibrated input range. 
We evaluate $B \in \{8, 12, 16\}$ on top of the W16A16 linear baseline.

At $B=16$, LUT replacement introduces little additional degradation for GPT-2 Small, Medium, and XL, and Qwen2.5-3B. 
However, the ratios reach about 32 for Qwen2.5-14B, 130 for Qwen3-30B-A3B, and $10^3$ for Qwen2.5-7B.
Thus, even B16 LUTs can fail catastrophically on larger models, despite preserving utility on smaller ones.

$\bullet$ \textit{Model and operator dependent precision requirements}. 
The same 16-bit LUT preserves utility for some models and leaves significant degradation in several large models. 
A fixed-precision LUT divides its calibrated input range
into uniform steps. For narrow ranges, even a smaller table can approximate the
operator well. For wide ranges, the same $B$ produces coarse steps whose error can compound through the residual stream. 
Therefore, it requires sufficient LUT resolution to approximate the original nonlinear operator over the inputs during inference.
The requirement depends on the model and operator, which motivates fine-grained diagnosis of utility bottlenecks. 
With our framework, we are able to isolate each LUT bit precision and evaluate whether targeting the bottleneck can help restore the lost utility.

\begin{table}[t]
  \centering
  \caption{
    PPL ratios relative to the W16A16 linear-only baseline:
    (a) full-LUT precision and
    (b) operator-wise diagnosis on WikiText-2. Complete results for all models are provided in Appendix~\ref{app:additional-utility}.
  }
  \label{tab:lut-results}

  \begingroup
  \scriptsize
  \setlength{\tabcolsep}{2pt}
  \renewcommand{\arraystretch}{1.15}

  \sbox{\FullLUTBox}{%
    \begin{tabular}{@{}lrrrrrr@{}}
      \toprule
      & \multicolumn{3}{c}{WikiText-2}
      & \multicolumn{3}{c}{C4} \\
      \cmidrule(lr){2-4}
      \cmidrule(l){5-7}
      Model & B8 & B12 & B16 & B8 & B12 & B16 \\
      \midrule
      GPT-2 S
        & 1866.2 & 1.011 & 1.000
        & 1439.4 & 1.018 & 1.002 \\
      GPT-2 M
        & 3213.6 & 1.002 & 1.001
        & 1465.4 & 1.000 & 0.999 \\
      Llama 8B
        & $>10^4$ & 1247.9 & 1126.8
        & $>10^4$ & 422.2 & 380.5 \\
      Qwen 14B
        & $>10^4$ & $>10^4$ & 32.5
        & $>10^4$ & $>10^4$ & 18.3 \\
      Qwen3 30B
        & $>10^4$ & 148.2 & 131.3
        & $>10^4$ & 78.7 & 120.7 \\
      \bottomrule
    \end{tabular}%
  }

  \setlength{\DiagnosisRowExtra}{0pt}
  \newcommand{\DiagnosisTable}{%
    \begin{tabular}{@{}lrrrr@{}}
      \toprule
      & \multicolumn{4}{c}{LUT configuration} \\
      \cmidrule(l){2-5}
      Model & Full B16 & RMS only & RMS exact & RMS B24 \\
      \midrule
      Llama 8B
        & 1126.8453 & 1126.7972 & 0.9993 & \textbf{1.0002}
        \\[\DiagnosisRowExtra]
      Qwen 7B
        & 1496.9087 & 1528.4988 & 0.9999 & \textbf{0.9987}
        \\[\DiagnosisRowExtra]
      Qwen 14B
        & 32.4609 & 32.6139 & 1.0002 & \textbf{0.9997}
        \\[\DiagnosisRowExtra]
      Qwen3 30B
        & 131.3352 & 1.0398 & 1.0034 & \textbf{1.0014}
        \\[\DiagnosisRowExtra]
      \bottomrule
    \end{tabular}%
  }

  \sbox{\DiagnosisBox}{\DiagnosisTable}
  \setlength{\DiagnosisRowExtra}{%
    \dimexpr
      \ht\FullLUTBox+\dp\FullLUTBox
      -\ht\DiagnosisBox-\dp\DiagnosisBox
    \relax
  }
  \divide\DiagnosisRowExtra by 4
  \sbox{\DiagnosisBox}{\DiagnosisTable}

  \resizebox{\textwidth}{!}{%
    \begin{tabular}{@{}c@{\hspace{14pt}}c@{}}
      \usebox{\FullLUTBox}
      & \usebox{\DiagnosisBox} \\[4pt]
      \textbf{(a) LUT precision sensitivity}
      & \textbf{(b) Targeted precision adjustment}
    \end{tabular}%
  }

  \vspace{4pt}
  \begin{minipage}{\textwidth}
    \scriptsize
    In (b), RMS only uses B16 RMS inverse-square-root LUTs
    with other nonlinear functions exact.
    RMS exact and RMS B24 use B16 for the other LUTs.
  \end{minipage}
  \endgroup
\end{table}

\paragraph{R3. RMSNorm inverse-square-root LUTs are a recurring utility bottleneck}
To identify the root cause of utility loss, we evaluate each model with a B16 LUT for one nonlinear operator. 
The rest operators are present in their original functions, which isolates the effect of a specific operator.

Table~\ref{tab:lut-results} (b) shows that RMS inverse square root is a shared bottleneck of four models. 
When we only use LUT for RMSNorm in Llama-3.1, Qwen2.5-7B, and 14B, the PPLs were very close to those of the full B16 pipeline. 
However, we observed that the single RMSNorm LUT setting does not impact Qwen3-30B like other models. 
Nevertheless, substituting the RMSNorm LUTs with their exact functions can bring the PPL back to near baseline, suggesting that interactions between operators lead to the utility loss.

$\bullet$ \textit{Root cause and increasing bit precision}.
We investigate the root cause of the bottleneck. At one RMSNorm site in Qwen2.5-7B, the calibrated input range extends from 0 to 65,461.31. A B16 LUT contains 65,536 entries, so the interval between table grid points is about 0.9989. 
According to our experiments, 66\% of the inputs are mapped to the grid point at zero. 
This implies that the grid of a B16 table is too coarse to represent most inputs faithfully. 
It leads to incorrect normalization and potentially disrupts the successive operations such as projection and residual computations. 
Increasing the LUT input precision to 24 bits improves the resolution. 
It reduces the grid spacing to 0.0039 over the same calibrated input range. 
None of the sampled inputs then maps to the grid point at zero. 
Instead, each input is represented by a nearby point on the finer grid.

$\bullet$ \textit{Why not use higher bit precision for all LUTs?} One could use high bit precision for all LUTs and it will likely lead to good utility. However, this will incur high overhead on both proving time and memory usage. Increasing the input precision from 16 to 24 bits expands the number of table entries by a factor of 256, which increases the memory usage from 256\,KiB to 64\,MiB per table with \texttt{int32} entries. The proving time is also proportional to the size of the table; see \S~\ref{sec:end-to-end-proving}. We provide fine-grained analysis for different non-linear operations and our results show that near-baseline utility can be achieved by introducing high-precision LUTs only for the RMSNorm layer.

\subsection{End-to-End Proving Costs}
\label{sec:end-to-end-proving}
\subsubsection{Floating-point vs. Quantized proving at Qwen-14B scale}
\label{subsubsec:fp-quantized}

In this section, we investigate the proving cost of ZK-friendly quantization. 
We begin with its cost advantages over floating-point proving.
Specifically, we target the projection dimensions of the largest dense model in our evaluation settings, Qwen2.5-14B. 
For floating-point arithmetic in ZK, we use the implementation of~\citet{ernstberger2025zero}, which was later used in~\citet{riasi2025zero} to prove ML computations. The protocol supports basic operations such as floating-point addition, multiplication, and division. Therefore, our comparison focuses on linear projections in the LLM as a lower-bound for the proving cost.  For quantized integer ZK proving, we measure the cost based on the zkML implementation of DeepProve~\citep{gailly2026deepprove}.

\begin{wraptable}{r}{0.49\textwidth}
  \vspace{-1.9\baselineskip}
  \centering
  \caption{Full-width projection proving times (seconds).}
  \label{tab:qwen-full-width}
  \footnotesize
  \setlength{\tabcolsep}{3pt}
  \renewcommand{\arraystretch}{1.08}
  \begin{tabular*}{\linewidth}{@{\extracolsep{\fill}}lrr@{}}
    \toprule
    Projection & Floating-point & Integer \\
       (each)        & (estimated)    & (measured) \\
    \midrule
    Q, O        & 12,158.78 & 11.877 \\
    K, V        &  2,431.76 &  4.603 \\
    FFN gate, up & 32,828.71 & 20.985 \\
    FFN down & 27,105.54 & 20.526 \\
    \midrule
    \textbf{Sum}
      & \textbf{121,944.05}
      & \textbf{95.457} \\
    \bottomrule
  \end{tabular*}
\end{wraptable}

Each Qwen-14B transformer block has seven projections, including Q, K, V, O, and the FFN gate, up, and down projections. Each projection is a matrix multiplication of the form $y=xW$, with an inner dimension of either 5,120 or 13,824. 
Since directly proving all output columns with the floating-point circuit is prohibitively expensive, we measure the proving time for eight output columns of each projection and extrapolate to the full projection width by scaling in proportion to the number of output columns. We then sum the estimated costs of the seven projections to obtain the estimated floating-point linear-projection proving cost. 

On the other hand, DeepProve allows us to directly prove the full-width projections with the W16A16 configuration.
Table~\ref{tab:qwen-full-width} compares these measured integer costs with the floating-point estimates.
We measure each distinct full-width projection shape once.
The integer total is the sum of these measured costs across the seven projections.
We put more detailed results with direct measurement of floating-point proving and RAM usage in Table~\ref{tab:qwen-linear-proving}, Appendix~\ref{sec:detailed_results}.

For one input token, the estimated floating-point proving cost of a single block is 33.87 hours, whereas the quantized integer proving time is 95.46 seconds. 
If we extend this to all 48 blocks, about 67 days will be taken for proving with floating-point numbers. 
This big gap occurs not only in proving time but also in peak RAM usage. 
Floating-point proving uses about 73--139 times as much peak RAM. 
This is because floating-point arithmetic requires additional constraints for exponent handling, normalization, and rounding.
Instead, ZK-friendly quantization uses integer arithmetic with fixed scales, avoiding these per-operation checks and handling rescaling through output requantization.
This experimental result justifies the importance of ZK-friendly quantization.

\subsubsection{Proving cost does not scale smoothly with quantization}
\paragraph{R4. End-to-end proving time with different precisions for small models}

In the previous section, we compared the proving costs of quantized integer and floating-point computation. Due to the high memory requirements of the current ZK protocol, full-model proof generation is not yet feasible at the largest model scale. We therefore evaluate end-to-end proving performance on smaller models (i.e., GPT-2  Small, Medium, and Large) across different integer precisions.  

\begin{figure}[htbp]
  \centering
  \includegraphics[width=.7\textwidth]{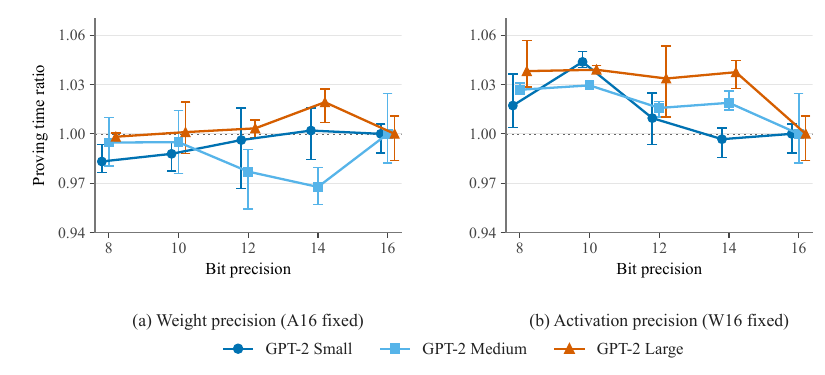}
  \vspace{-15pt}
  \caption{Proving time ratio relative to the W16A16 baseline for each model.}
  \label{fig:wa_proving}
\end{figure}

Figure~\ref{fig:wa_proving} shows the proving time for GPT-2 Small, Medium, and Large, normalized by each model's W16A16 mean. The left panel examines the weight precision sensitivity and the right panel is for the activation sensitivity. 
We run each configuration three times. 

We observe that proving cost does not strongly depend on the bit precision in this range. This is because as long as the computations do not lead to overflows in $\mathbb{F}_p$, different precisions result in the same number of arithmetic operations and lookups in the ZK protocol. That is, $|\mathcal{R}_{Q,p}|$ is roughly the same for these quantization schemes. Still, bit precision affects the integer representation and range of values, leading to different requantizations and range decompositions. However, in the experiments, we observe that these precision-dependent components are too small to materially change total prover time. We put more detailed results in Appendix~\ref{sec:context-proving-cost}. Our result suggests that for small models that have already been measured in prior work~\citep{qu2025zkgpt,gailly2026deepprove}, one could use higher precisions such as W16A16 for better utilities, with minimal overhead on the proving cost. 

However, the result above does not mean we can always increase the precision with no cost. 
At higher precision, products and accumulated sums may no longer fit within the range supported by the underlying finite field.
For example, multiplying two integers wider than 16 bits can exceed the capacity of a small field, such as the 31-bit BabyBear field $p=15\cdot2^{27}+1$ or the Mersenne prime $p=2^{31}-1$, which are commonly used in many ZK protocols due to fast arithmetic over these fields~\citep{plonky3,risczero}. Once the overflow occurs, each integer value needs to be represented by multiple field elements, which will significantly increase $|\mathcal{R}_{Q,p}|$, and thus the proving cost. Therefore, the proving cost does not scale smoothly with the bit precision, and our work provides the first framework to select the minimum bit precision with good utility for different models for future zkML designs.

\paragraph{R5. End-to-end proving cost with different LUT sizes}

We measure the end-to-end proving cost with different LUT sizes as well.
We vary both the GeLU and softmax exponential LUT bit precision between 8 and 24 bits in GPT-2 Small. We observe similar phenomena of non-smooth scaling as well. Reducing both LUTs from B16 to B8 shrinks each table by \(256\times\), yet the mean proving time remains essentially unchanged (102.65 vs. 104.75 s). However, increasing the LUTs to B24 makes the proving time 2.6$\times$ slower (265.54 s). 
More configurations are reported in Table~\ref{tab:lut-cost} in Appendix~\ref{sec:detailed_results}. 

The reason is different from the bit precision. Recall that the proving cost of lookup arguments is proportional to both the size of the LUT and the number of lookup queries. In this experiment, for the same model the number of direct lookup queries remains fixed at 0.823 million; reducing table precision changes table size, but not the number of lookups, which is the same as the number of GeLU and exponential evaluations. Reducing the LUT size from B16 to B8 decreases part of the proving time related to the LUT from 1.325 to 0.483 seconds, but this is not the dominating cost in end-to-end proving. However, increasing the LUT size from B16 to B24 makes the table size the new bottleneck (16 million entries), thus significantly increasing the prover time. These results justify the necessity of our fine-grained utility analysis in \S~\ref{subsec:broad-utility}, where we only need a B24 LUT for the RMSNorm while keeping other LUTs at B16.

\section{Conclusion}
\label{sec:conclusion}
We introduce a formalization of ZK-friendly quantization and a scalable framework for studying its effects on LLM utility and ZK proving cost. We evaluate nine language models across weight, activation, and nonlinear lookup precisions. Our experiments identify RMSNorm inverse-square-root approximation as a recurring utility bottleneck in several models at 7B scale and above, and show that targeted increases in LUT precision can recover near-baseline utility without uniformly increasing the precision of all nonlinear operators. We further show that end-to-end proving cost does not necessarily scale linearly with quantization bit-width or LUT size. 
These findings suggest three design principles for ZK-LLM quantization: preserve sufficient activation precision, allocate nonlinear precision on an operator-specific basis, and reduce numerical precision only when it meaningfully simplifies the underlying proof relation. Together, these principles motivate heterogeneous, operator-aware quantization for balancing model fidelity with proving efficiency.

\subsubsection*{Acknowledgments}
This work is partially supported by the National Science Foundation (NSF) under Grant No. 2453149, 2613388 and Amazon research award. Any opinions, findings, and conclusions or recommendations expressed in this material are those of the author(s) and do not necessarily reflect the views of these institutes.

\bibliography{iclr2027_conference}
\bibliographystyle{iclr2027_conference}

\clearpage
\appendix

\section{Simulated Integer Execution}
\label{sec:qdq_simulation}

Our framework uses quantize-dequantize (QDQ) simulation to evaluate quantization configurations with existing PyTorch operations and GPU acceleration. 
This avoids implementing dedicated integer kernels for every bit-width, such as INT12, and makes large-model evaluation practical.
It raises one question: how closely does their output reproduce the corresponding integer computation? 
We answer this question first algebraically and then experimentally. 
We separate the error introduced by quantization from the additional error introduced by its QDQ simulation. 

\paragraph{Algebraic comparison}
QDQ first rounds and clips each tensor to its bounded-integer representation, then dequantizes it before the PyTorch operation. 
Dequantization restores the scale of the values, but not the information lost through rounding and clipping.
Let $X \in \mathbb{R}^{M\times K}$ and $W \in \mathbb{R}^{K\times N}$ have fixed per-tensor scales $s_x$ and $s_w$. 
For either tensor $Z$ with scale $s_z$, we define signed $q$-bit quantization as
\begin{equation}
    Q_Z =
    \operatorname{clip}_{[-2^{q-1},\,2^{q-1}-1]}
    \!\left(
        \operatorname{round}\!\left(\frac{Z}{s_z}\right)
    \right).
\end{equation}
The QDQ and integer paths use the same quantized tensors but apply their
scales at different stages:
\begin{align}
    Y_{\mathrm{int}} &= (Q_X Q_W)s_xs_w
        && \text{(integer multiplication, then rescale)},
    \\
    Y_{\mathrm{QDQ}} &= (s_x Q_X)(s_w Q_W)
        && \text{(dequantize, then multiply)}.
\end{align}

The above expressions are identical under exact arithmetic.
On a computer, however, dequantization and matrix multiplication round the number, so algebraic equivalence does not promise numerical agreement.

\paragraph{Experimental validation}
We examine the numerical errors introduced by quantization and QDQ in actual computation. 
Using nine projections in total from GPT-2 Small, we compute three outputs: the unquantized product $Y_{\mathrm{FP}}=XW$, the integer-accumulation reference $Y_{\mathrm{int}}$, and the QDQ output $Y_{\mathrm{QDQ}}$.
From the configurations used in our experiments, we select W8A8, W12A12, and W16A16, testing both the standard scale used in the main experiments and the power-of-two (PoT) scale. 
PoT provides a useful comparison because binary floating-point arithmetic can apply power-of-two scaling exactly when the scaled values are representable.
We execute the QDQ projections in both FP32 and BF16.

We define two errors to distinguish the effect of quantization from the additional error introduced by its simulation.
The quantization error, $E_{\mathrm{quant}}$, measures how much quantization changes the original floating-point computation.
Both the integer and QDQ paths use quantized operands, so this quantization effect is present in both paths.
We isolate it by comparing $Y_{\mathrm{int}}$ with $Y_{\mathrm{FP}}$.
The simulation error, $E_{\mathrm{sim}}$, instead measures how closely QDQ reproduces the integer reference.
It captures the additional numerical error introduced by floating-point dequantization, dtype conversion, and matrix multiplication.
Thus, even when $E_{\mathrm{quant}}$ is large, a sufficiently small $E_{\mathrm{sim}}$ implies that QDQ faithfully reproduces the quantized computation. 
To summarize the errors across all nine projections, we flatten and concatenate their outputs into one vector for each computation path.

\begin{equation}
E_{\mathrm{quant}}
=
\frac{
    \left\|Y_{\mathrm{int}}-Y_{\mathrm{FP}}\right\|_2
}{
    \left\|Y_{\mathrm{FP}}\right\|_2
},
\qquad
E_{\mathrm{sim}}
=
\frac{
    \left\|Y_{\mathrm{QDQ}}-Y_{\mathrm{int}}\right\|_2
}{
    \left\|Y_{\mathrm{int}}\right\|_2
}.
\end{equation}

\begin{table}[ht]
    \centering
    \caption{Numerical validation of QDQ on actual GPT-2 Small projections.
    All values are relative errors, not percentages.}
    \label{tab:qdq-validation}
    \small
    \setlength{\tabcolsep}{5pt}
    \renewcommand{\arraystretch}{1.08}
    \begin{tabular*}{\textwidth}{@{\extracolsep{\fill}}llrrr@{}}
        \toprule
        & & & \multicolumn{2}{c}{Simulation error $E_{\mathrm{sim}}$} \\
        \cmidrule(lr){4-5}
        Scale & Precision & Quantization error $E_{\mathrm{quant}}$
            & FP32 & BF16 \\
        \midrule
        Standard & W8A8   & $6.8022\times10^{-1}$ & $2.2056\times10^{-7}$ & $2.4294\times10^{-3}$ \\
                & W12A12 & $1.2246\times10^{-2}$ & $3.1378\times10^{-7}$ & $2.2690\times10^{-3}$ \\
                & W16A16 & $7.2081\times10^{-4}$ & $3.1691\times10^{-7}$ & $1.7096\times10^{-3}$ \\
        \midrule
        PoT     & W8A8   & $6.7154\times10^{-1}$ & $0$ & $1.6644\times10^{-3}$ \\
                & W12A12 & $2.2929\times10^{-2}$ & $0$ & $1.6218\times10^{-3}$ \\
                & W16A16 & $1.0408\times10^{-3}$ & $1.1761\times10^{-7}$ & $1.6343\times10^{-3}$ \\
        \bottomrule
    \end{tabular*}
\end{table}

\paragraph{Results.}
Table~\ref{tab:qdq-validation} presents the numerical validation results. 
As expected, increasing quantization precision reduces the error from the unquantized floating-point result.
Our main interest is the simulation error. 
In particular, the PoT configurations produce identical outputs at 8 and 12 bits in this experiment.
This is because power-of-two scaling can be applied exactly to representable binary floating-point values, removing a source of rounding error.
With BF16 execution, the simulation error is generally larger than with FP32, but still remains small.
BF16 inherently has fewer significant bits than FP32, so storing the dequantized operands and outputs in BF16 introduces additional rounding. 
For some cases that require more numerical equivalence, FP32 execution with PoT scaling can be a useful option. 
This choice, however, comes with the higher memory requirements of FP32 execution.

\section{Lookup Tables}
\label{sec:lookup}
In our framework, the key mechanism for replacing non-linear operations is a LUT, a fixed
table of precomputed input-output pairs for a target function. For a non-linear operator $g$, we first choose an input range by calibration or an operator-specific predefined bound. 
Once we fix the input range, we discretize it into a uniform grid
with $2^B$ entries, where $B$ denotes the LUT input precision.
We then evaluate $g$ at each grid point and quantize the outputs
to 24-bit integers stored in \texttt{int32}.
During evaluation, each runtime input is rounded to the nearest grid and replaced by the corresponding output in the table. 
The LUT for softmax exponential, for example, has an input range of $[-20, 0]$. 
If the input is out of the range, then we clip it to the nearest endpoint. Our simulation collects aggregate clipping rates and lookup counts per model.

The PyTorch implementation follows the same rule in general, while the exact implementation for each non-linear operator is different. 
This is because the place where the operator appears and the way it is called are different. 
For example, we replace the forward computation of GeLU and SiLU modules with LUT evaluation while preserving the existing modules.
For normalization, we modify the forward computation to replace only the inverse-square-root function with a LUT, preserving the mean and variance computation in LayerNorm and the mean-square computation in RMSNorm.
For attention softmax, we temporarily patch the softmax evaluation path so that the exponential and reciprocal are computed through LUTs.
In particular, for Qwen3, LUT coverage also includes Q/K RMSNorm and the exponential and reciprocal operations used in MoE routing.
In terms of memory, an \texttt{int32} B16 table requires 256\,KiB, whereas a B24 table requires 64\,MiB, both much smaller than the model weights in our experiments.
Additionally, lookup is implemented as batched tensor indexing rather than element-wise Python lookup, avoiding per-element Python overhead.

\section{Detailed Framework Implementation}
\label{sec:detailed_implementation}
Our framework implements ZK-friendly quantization as a config-driven PyTorch and Hugging Face pipeline. Given a model and a quantization configuration, it patches the transformer forward pass at the operation level:
linear tensor operations use quantized weights and activations with quantize-dequantize modules, while selected nonlinear functions are replaced by LUT modules (See Appendix~\ref{sec:lookup} for a more detailed explanation of non-linear operations). The configuration specifies the design choices that affect both ZK compatibility and model utility, including weight and activation bit-widths, scale type, quantization symmetry, activation calibration, and LUT bit-width.

\paragraph{Linear operations.}
All matrix multiplications in the transformer produce unbounded floating-point intermediate results that must be converted to bounded integers for ZK circuit compatibility. This category includes every \texttt{nn.Linear} (or \texttt{Conv1D} in GPT-2) layer in the
architecture, such as the QKV projections, the output projection, the MLP layers, and the final language-modeling head. 
The language-modeling head uses quantized weights but does not apply an additional A-bit requantization to its output logits.
Attention score computation $QK^\top$ and context aggregation $\mathrm{softmax}(S)V$ remain native tensor operations; the evaluated policy does not insert a separate QDQ operation after each of these matrix products. 
The evaluation policy also covers embedding weights and residual outputs, whose precision is fixed at 24 bits in the main utility experiments, as well as the expert and router projections in Qwen3.
Given this operation-level classification, the framework implements each configuration by patching the corresponding PyTorch modules in place.

\paragraph{Scales.}
For each quantized tensor, the configuration determines how its scale is chosen. We use one scale per weight tensor in all experiments. 
A standard scale uses the floating-point range of the tensor directly. 
The main utility experiments use the scale $s=2a/(2^b-1)$, whereas the standard-scale ablation uses $s=a/(2^{b-1}-1)$, with $a$ meaning the maximum absolute value and $b$ the target bit-width.
A power-of-two scale is computed as
$s=2^{\lceil\log_2(a/2^{b-1})\rceil}$.
The reason a power-of-two scale is ZK-friendly is that it can be compiled as a fixed shift-like integer operation or a multiplication by a public constant. 
The $c_1/c_2$ scale uses an integer-ratio rescaling rule, where $c_1$ and $c_2$ are fixed integers as in zkGPT~\citep{qu2025zkgpt}. 
It can avoid floating-point arithmetic by using integer multiplication and division by a constant.

\paragraph{Activation calibration.}
Activation calibration is needed because activations are runtime values. 
We calibrate activation ranges on eight fixed WikiText-2 training windows with context length 512 and stride 256, before weight quantization. 
For each static configuration, scales are determined from these ranges and reused across evaluations, including C4.
We use one static scale per designated activation tensor, rather than one scale for an entire transformer layer.
For a linear layer $y=xW+b$, the weights $W$ and bias $b$ are fixed and can be quantized offline, but the output activation $y$ depends on the input prompt. The output $y$ also becomes the input of the next layer. 
Static-max calibration fixes activation scales before inference~\citep{xie2025zkpytorch, xiao2023smoothquant}. We run calibration prompts once, record a maximum activation range at each designated site, and reuse that fixed range during evaluation. 
Dynamic activation quantization instead computes the scale from the current input, which incurs a lot of overhead.

\paragraph{Compute environment.}
The main model utility experiments were run with two Intel Xeon Gold 6526Y CPUs, 251 GiB RAM, and two NVIDIA H100 NVL GPUs. The PyTorch environment used version 2.10.0 with CUDA 12.6.

\section{Additional Utility Results}
\label{app:additional-utility}

\begin{figure}[htbp]
  \centering
  \includegraphics[width=.8\textwidth]{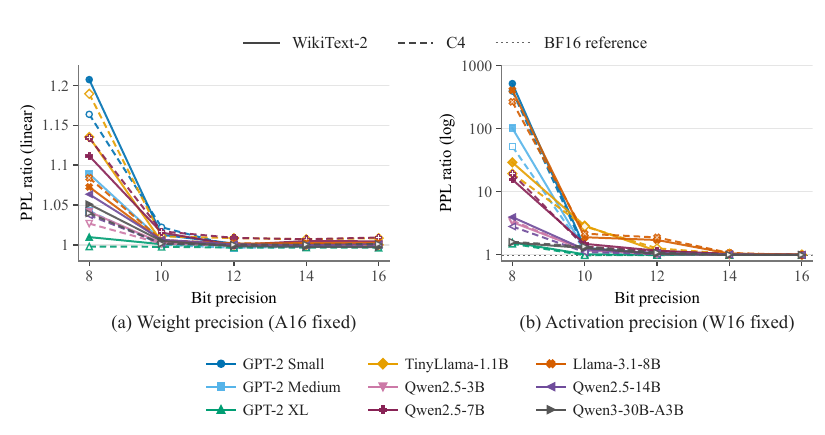}
  \vspace{-15pt}
  \caption{Perplexity ratio under linear quantization to BF16 baseline.}
  \label{fig:linear-quant-all}
\end{figure}

\begin{table}[ht]
  \centering
  \caption{
    Full-LUT PPL ratios relative to the W16A16 linear-only baseline.
  }
  \label{tab:full-lut-all}

  \begingroup
  \scriptsize    
  \setlength{\tabcolsep}{3pt}
  \renewcommand{\arraystretch}{1.08}
  \resizebox{0.8\textwidth}{!}{%
    \begin{tabular}{@{}lcrrrrrrrrr@{}}
      \toprule
      \multirow{2}{*}{Dataset}
      & \multirow{2}{*}{$B$}
      & \multicolumn{3}{c}{GPT-2}
      & \multicolumn{1}{c}{TinyLlama}
      & \multicolumn{1}{c}{Llama-3.1}
      & \multicolumn{3}{c}{Qwen2.5}
      & \multicolumn{1}{c}{Qwen3} \\
      \cmidrule(lr){3-5}
      \cmidrule(lr){6-6}
      \cmidrule(lr){7-7}
      \cmidrule(lr){8-10}
      \cmidrule(l){11-11}
      & & Small & Medium & XL
      & 1.1B & 8B & 3B & 7B & 14B & 30B-A3B \\
      \midrule

      \multirow{3}{*}{WikiText-2}
      & 8
      & 1866.2 & 3213.6 & 704.3
      & $>10^4$ & $>10^4$ & $>10^4$
      & $>10^4$ & $>10^4$ & $>10^4$ \\
      & 12
      & 1.011 & 1.002 & 1.002
      & 2356.3 & 1247.9 & 125.5
      & 5254.5 & $>10^4$ & 148.2 \\
      & 16
      & 1.000 & 1.001 & 1.000
      & 3.073 & 1126.8 & 1.000
      & 1496.9 & 32.5 & 131.3 \\

      \midrule

      \multirow{3}{*}{C4}
      & 8
      & 1439.4 & 1465.4 & 316.7
      & $>10^4$ & $>10^4$ & $>10^4$
      & $>10^4$ & $>10^4$ & $>10^4$ \\
      & 12
      & 1.018 & 1.000 & 0.989
      & 2258.8 & 422.2 & 78.8
      & 3472.9 & $>10^4$ & 78.7 \\
      & 16
      & 1.002 & 0.999 & 1.000
      & 3.126 & 380.5 & 0.999
      & 769.6 & 18.3 & 120.7 \\

      \bottomrule
    \end{tabular}%
  }
  \endgroup
\end{table}

\begin{table}[ht]
  \centering
  \caption{
    PPL ratios relative to the W16A16 linear-only baseline
    with targeted precision adjustment.
  }
  \label{tab:operator-diagnosis-all}

  \setlength{\tabcolsep}{3pt}
  \renewcommand{\arraystretch}{1.08}

  \resizebox{0.8\textwidth}{!}{%
    \begin{tabular}{@{}lrrrr@{}}
      \toprule
      LUT configuration
        & Llama-3.1-8B
        & Qwen2.5-7B
        & Qwen2.5-14B
        & Qwen3-30B-A3B \\
      \midrule
      Full B16
        & 1126.8453 & 1496.9087 & 32.4609 & 131.3352 \\
      RMS B16 only; others exact
        & 1126.7972 & 1528.4988 & 32.6139 & 1.0398 \\
      Full B16; RMS exact
        & 0.9993 & 0.9999 & 1.0002 & 1.0034 \\
      \addlinespace[2pt]
      Full B16; RMS B24
        & \textbf{1.0002}
        & \textbf{0.9987}
        & \textbf{0.9997}
        & \textbf{1.0014} \\
      \bottomrule
    \end{tabular}%
  }
\end{table}

Figure~\ref{fig:linear-quant-all} extends the linear-quantization comparison to all nine models on WikiText-2 and C4.
Activation quantization generally causes greater utility loss than weight quantization, although the sensitivity varies across models.
Table~\ref{tab:full-lut-all} reports the additional effect of LUT approximation relative to the W16A16 linear-only baseline.
B8 LUTs cause severe degradation across all models, while increasing precision to B16 preserves utility for some models but remains insufficient for others.

Table~\ref{tab:operator-diagnosis-all} examines four affected models on WikiText-2.
RMS inverse-square-root LUTs alone reproduce much of the full-B16 degradation in Llama-3.1-8B and Qwen2.5-7B/14B.
For Qwen3-30B-A3B, the degradation instead emerges when these LUTs are combined with the other approximations.
Replacing only the RMS LUTs with exact functions brings all four models close to their linear-only baselines.
Increasing their input precision to B24 achieves a similar result.

\section{Additional Proving Cost Results}
\label{sec:context-proving-cost}

\paragraph{Weight and activation precision}
Table~\ref{tab:wa-proving-cost} reports the absolute proving times underlying Figure~\ref{fig:wa_proving}.
We evaluate GPT-2 Small, Medium, and Large at a context length of 16 tokens, with batch size one and 32 CPU threads.
We report the mean, minimum, and maximum proving times.
The reported time excludes graph preparation, inference,
and verification.

\begin{table}[ht]
  \centering
  \caption{Proving time under different W/A precisions (seconds).}
  \label{tab:wa-proving-cost}
  \small
  \setlength{\tabcolsep}{3pt}
  \renewcommand{\arraystretch}{1.08}
  \begin{tabular*}{\textwidth}{
    @{\extracolsep{\fill}}l*{9}{r}@{}
  }
    \toprule
    & \multicolumn{3}{c}{GPT-2 Small}
    & \multicolumn{3}{c}{GPT-2 Medium}
    & \multicolumn{3}{c}{GPT-2 Large} \\
    \cmidrule(lr){2-4}
    \cmidrule(lr){5-7}
    \cmidrule(lr){8-10}
    Configuration
      & Mean & Min & Max
      & Mean & Min & Max
      & Mean & Min & Max \\
    \midrule
    W16A16
      & 104.474 & 103.257 & 105.087
      & 222.355 & 218.421 & 227.839
      & 525.555 & 517.006 & 531.283 \\
    \midrule
    W8A16
      & 102.718 & 102.015 & 103.790
      & 221.179 & 218.017 & 224.541
      & 524.634 & 523.111 & 525.836 \\
    W10A16
      & 103.205 & 102.106 & 103.921
      & 221.231 & 216.977 & 225.548
      & 526.105 & 519.276 & 535.732 \\
    W12A16
      & 104.079 & 101.009 & 106.103
      & 217.248 & 212.212 & 220.218
      & 527.309 & 524.410 & 530.007 \\
    W14A16
      & 104.683 & 102.837 & 106.115
      & 215.175 & 212.812 & 217.813
      & 535.667 & 529.128 & 539.949 \\
    \addlinespace[3pt]
    W16A8
      & 106.272 & 104.882 & 108.290
      & 228.313 & 227.716 & 229.210
      & 545.609 & 540.546 & 555.396 \\
    W16A10
      & 109.046 & 108.687 & 109.711
      & 228.912 & 228.600 & 229.375
      & 546.040 & 545.187 & 547.397 \\
    W16A12
      & 105.473 & 103.788 & 107.059
      & 225.853 & 224.640 & 226.734
      & 543.240 & 530.995 & 553.663 \\
    W16A14
      & 104.135 & 102.978 & 104.835
      & 226.552 & 225.606 & 228.140
      & 545.261 & 540.023 & 549.084 \\
    \bottomrule
  \end{tabular*}
\end{table}

\paragraph{Context length}
We extend the W/A precision experiment by measuring GPT-2 Small and Medium at context lengths of 4, 8, 16, and 32 tokens, using W16A16 and W12A16.
We use the same quantization policies and DeepProve backend as in the main experiment (\S~\ref{sec:end-to-end-proving}), with batch size one and 32 CPU threads.
For each setting, we prove the full prefill computation and report one formal measurement.
The reported time excludes graph preparation, inference, and verification.

\begin{table}[t]
  \centering
  \caption{Proving time across context lengths (seconds).}
  \label{tab:context-proving-cost}
  \small
  \setlength{\tabcolsep}{5pt}
  \renewcommand{\arraystretch}{1.08}
  \begin{tabular*}{\textwidth}{@{\extracolsep{\fill}}crrrr@{}}
    \toprule
    & \multicolumn{2}{c}{GPT-2 Small}
    & \multicolumn{2}{c}{GPT-2 Medium} \\
    \cmidrule(lr){2-3}\cmidrule(lr){4-5}
    Context length & W16A16 & W12A16 & W16A16 & W12A16 \\
    \midrule
     4 &  73.116 &  74.093 & 158.405 & 150.433 \\
     8 &  87.156 &  85.499 & 183.162 & 179.619 \\
    16 & 103.666 & 100.756 & 225.444 & 218.262 \\
    32 & 125.902 & 130.727 & 274.920 & 274.536 \\
    \bottomrule
  \end{tabular*}
\end{table}

Table~\ref{tab:context-proving-cost} shows that proving time increases consistently with context length in both models.
Increasing the context from 4 to 32 tokens raises proving time by about 1.8 times across the four settings.
The increase with context follows a change in the amount of computation:
longer inputs produce more activation elements and larger attention matrices whose relations must be proved.
The recorded number of intermediate elements increases by approximately 8.4 times from context 4 to 32, whereas reducing weight precision leaves both the graph node count and intermediate tensor sizes unchanged at each context.
In conclusion, we observe that context length has a much larger effect on proving time than the change of weight precision.

\section{Behavioral Fidelity Experimentation}
\label{sec:behavioral_fidelity}
Perplexity is a useful scalar measure of language-modeling utility, but it is not sufficient for ZK-LLM governance, where the proved computation may depend on fine-grained output-distribution behavior. 
Auditing model behavior, verifying token-level claims, and proving membership-inference statistics require quantization to preserve token rankings, low-probability regions, and downstream decisions, not only average likelihood. 
We therefore evaluate W16A16 and W12A12 with the model-specific LUT configurations selected in our utility study (\S~\ref{subsec:broad-utility}), alongside a W16A16 linear-only baseline.

$\bullet$ \textit{Logit-level fidelity}.

Table~\ref{tab:logit_fidelity_exact} compares BF16 and quantized next-token distributions on WikiText-2 and C4 for the largest five models in our suite.
Each configuration is evaluated once on the same 8,192 scored tokens per model and dataset.
For each token position, we compute KL divergence from the BF16 distribution to the quantized distribution and the fraction of shared tokens in their Top-5 predictions.
We also measure bottom-20\% overlap: each model selects the 20\% of evaluated positions with the lowest probabilities assigned to the actual target tokens, and we report the intersection divided by the selected set size.

The full-pipeline configurations use B16 LUTs throughout Qwen2.5-3B.
For the other four models, block and final RMSNorm inverse-square-root LUTs use B24, while the remaining LUTs use B16.
Qwen3's Q/K RMSNorm and router LUTs also remain at B16.
The same LUT settings and calibrated ranges are used for W16A16 and W12A12.

W16A16 with these LUT configurations consistently preserves token-level behavior. 
It maintains low KL divergence and high overlap in both the top-ranked and low-probability token regions across datasets and models. 
This matters for ZK-LLM governance applications because the proof may certify a computation over token probabilities, not only a final generated token. 
By contrast, W12A12 with the same LUT settings has lower fidelity across all five models. 
This confirms that perplexity alone is not enough to certify behavioral fidelity: a configuration can appear acceptable under scalar likelihood while still changing distributional structure. 

\begin{table}[t]
  \centering
  \caption{Logit-level fidelity relative to the BF16 reference.}
  \label{tab:logit_fidelity_exact}
  \small
  \setlength{\tabcolsep}{3pt}
  \renewcommand{\arraystretch}{1.08}
  \begin{tabular*}{\textwidth}{
    @{\extracolsep{\fill}}llrrrrrr@{}
  }
    \toprule
    & & \multicolumn{3}{c}{WikiText-2}
      & \multicolumn{3}{c}{C4} \\
    \cmidrule(lr){3-5}
    \cmidrule(lr){6-8}
    Model & Configuration
      & KL $\downarrow$ & Top-5 $\uparrow$ & Bottom-20\% $\uparrow$
      & KL $\downarrow$ & Top-5 $\uparrow$ & Bottom-20\% $\uparrow$ \\
    \midrule

    Qwen2.5-3B
      & W16A16, linear
      & 0.001847 & 0.9726 & 0.9872
      & 0.001570 & 0.9769 & 0.9884 \\
      & W16A16, full
      & 0.003710 & 0.9612 & 0.9835
      & 0.002999 & 0.9695 & 0.9853 \\
      & W12A12, full
      & 0.040696 & 0.8842 & 0.9505
      & 0.030019 & 0.9032 & 0.9536 \\

    \addlinespace[4pt]
    Qwen2.5-7B
      & W16A16, linear
      & 0.015128 & 0.9354 & 0.9719
      & 0.019180 & 0.9307 & 0.9670 \\
      & W16A16, full
      & 0.014987 & 0.9363 & 0.9701
      & 0.013916 & 0.9395 & 0.9744 \\
      & W12A12, full
      & 0.214690 & 0.7761 & 0.8675
      & 0.155822 & 0.8123 & 0.8858 \\

    \addlinespace[4pt]
    Llama-3.1-8B
      & W16A16, linear
      & 0.003787 & 0.9617 & 0.9823
      & 0.003608 & 0.9654 & 0.9841 \\
      & W16A16, full
      & 0.004143 & 0.9625 & 0.9829
      & 0.006160 & 0.9613 & 0.9829 \\
      & W12A12, full
      & 0.551584 & 0.6525 & 0.7595
      & 0.764554 & 0.6052 & 0.7314 \\

    \addlinespace[4pt]
    Qwen2.5-14B
      & W16A16, linear
      & 0.004596 & 0.9577 & 0.9811
      & 0.002593 & 0.9706 & 0.9829 \\
      & W16A16, full
      & 0.004642 & 0.9584 & 0.9786
      & 0.002723 & 0.9700 & 0.9841 \\
      & W12A12, full
      & 0.099876 & 0.8619 & 0.9115
      & 0.039257 & 0.9035 & 0.9530 \\

    \addlinespace[4pt]
    Qwen3-30B-A3B
      & W16A16, linear
      & 0.005125 & 0.9600 & 0.9853
      & 0.003375 & 0.9661 & 0.9853 \\
      & W16A16, full
      & 0.005489 & 0.9604 & 0.9799
      & 0.003477 & 0.9658 & 0.9847 \\
      & W12A12, full
      & 0.105282 & 0.8492 & 0.9170
      & 0.097227 & 0.8510 & 0.9145 \\

    \bottomrule
  \end{tabular*}

  \vspace{3pt}
  \begin{minipage}{\textwidth}
    \footnotesize
    Linear: nonlinear functions remain exact.
    Full: the model-specific B16/B24 LUT configuration described
    in the text, held fixed across W/A settings.
  \end{minipage}
\end{table}

$\bullet$ \textit{Downstream task accuracy}.
We evaluate two multiple-choice
benchmarks: ARC-easy~\citep{clark2018think} for basic reasoning and
MMLU~\citep{hendrycks2020measuring} for broad knowledge, using the lm-evaluation-harness~\citep{gao2021framework} MCQ scoring protocol.
These tasks test whether the quantized model preserves end-task decisions, rather than only pretraining likelihood or token-distribution similarity.
We evaluate all 2,376 ARC-easy test questions with 0-shot prompts and all 14,042 MMLU test questions across 57 subjects with 5-shot prompts.

\begin{figure}[htbp]
  \centering
  \includegraphics[width=.9\textwidth]{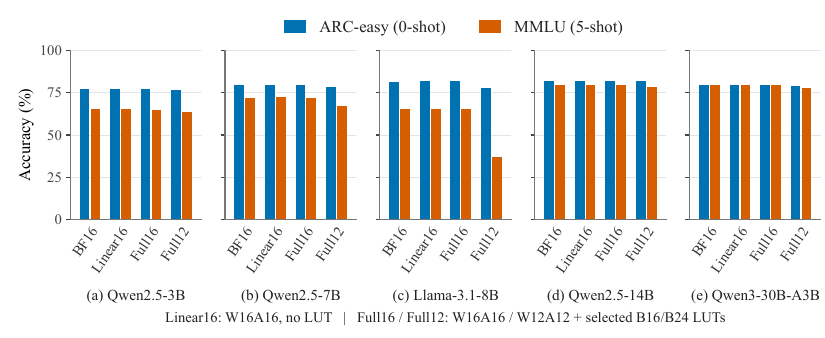}
  \vspace{-12pt}
  \caption{Downstream task accuracy on ARC-easy and MMLU under different configurations.}
  \label{fig:downstream_accuracy}
\end{figure}

Figure~\ref{fig:downstream_accuracy} describes the accuracy results under different quantization configurations. 
Both Full16 and Full12 settings are from our utility study: B24 for block and final RMSNorm lookups with B16 for the rest. 
The result suggests that the strongest ZK-friendly configuration (Full 16) preserves downstream behavior as well as perplexity and logit-level fidelity. 
While Full 12 is more model-dependent. In particular, it shows larger loss in Llama-3.1-8B. 
Hence, the results present that our framework can evaluate quantization configurations not only through perplexity and logit-level fidelity, but also through downstream tasks that can reflect practical use of LLMs.

\section{Quantization Design-Choice Ablations}
\label{sec:design_ablation}

We examine how quantization design choices affect utility at W12A12 and W16A16 across nine models on WikiText-2 and C4.
The baseline uses static per-tensor signed-symmetric quantization with non-power-of-two scales.
Starting from this baseline, we separately substitute power-of-two scales, conventional symmetric scales, bounded integer-ratio scales, asymmetric quantization, or dynamic per-token activation quantization.
The dynamic setting uses the baseline weight quantization.
Across all settings, embedding and residual quantizers remain fixed at 24 bits, and nonlinear functions are evaluated without LUT approximation.
Each configuration is evaluated once on 32 windows per dataset, with a context length of 512 and 8,192 scored tokens in total.

\begin{table}[t]
  \centering
  \caption{Design-choice ablations: perplexity with exact nonlinear functions.}
  \label{tab:full_ablation}
  \scriptsize
  \setlength{\tabcolsep}{2.5pt}
  \renewcommand{\arraystretch}{1.08}
  \resizebox{\textwidth}{!}{%
  \begin{tabular}{@{}llrrrrrr@{\hspace{8pt}}rrrrrr@{}}
    \toprule
    & & \multicolumn{6}{c}{W12A12} & \multicolumn{6}{c}{W16A16} \\
    \cmidrule(lr){3-8}\cmidrule(lr){9-14}
    Model & Dataset & Base & PoT & Conv. & $c_1/c_2$ & Asym & Dyn & Base & PoT & Conv. & $c_1/c_2$ & Asym & Dyn \\
    \midrule
    GPT-2 Small & WikiText-2 & 27.05 & 28.53 & 27.18 & 26.99 & 26.60 & 26.45 & 26.35 & 26.32 & 26.41 & 26.28 & 26.33 & 26.37 \\
     & C4 & 31.99 & 33.43 & 32.00 & 31.99 & 31.31 & 31.42 & 31.20 & 31.27 & 31.30 & 31.26 & 31.14 & 31.31 \\
    \addlinespace[2pt]
    GPT-2 Medium & WikiText-2 & 19.77 & 19.93 & 19.80 & 19.79 & 19.57 & 19.35 & 19.31 & 19.34 & 19.34 & 19.32 & 19.36 & 19.29 \\
     & C4 & 24.99 & 25.17 & 24.92 & 25.06 & 25.02 & 24.85 & 24.82 & 24.80 & 24.86 & 24.88 & 24.84 & 24.75 \\
    \addlinespace[2pt]
    GPT-2 XL & WikiText-2 & 15.33 & 15.27 & 15.33 & 15.32 & 15.30 & 15.30 & 15.31 & 15.31 & 15.31 & 15.31 & 15.31 & 15.30 \\
     & C4 & 19.70 & 19.70 & 19.68 & 19.69 & 19.68 & 19.75 & 19.70 & 19.76 & 19.71 & 19.71 & 19.68 & 19.77 \\
    \addlinespace[2pt]
    TinyLlama-1.1B & WikiText-2 & 9.05 & 9.93 & 9.05 & 9.08 & 8.84 & 7.91 & 7.91 & 7.93 & 7.91 & 7.91 & 7.92 & 7.91 \\
     & C4 & 15.32 & 16.90 & 15.26 & 15.36 & 14.64 & 12.15 & 12.16 & 12.16 & 12.17 & 12.17 & 12.08 & 12.13 \\
    \addlinespace[2pt]
    Qwen2.5-3B & WikiText-2 & 7.79 & 7.91 & 7.81 & 7.80 & 7.73 & 7.57 & 7.57 & 7.58 & 7.57 & 7.57 & 7.57 & 7.57 \\
     & C4 & 13.72 & 13.94 & 13.68 & 13.69 & 13.58 & 13.40 & 13.39 & 13.39 & 13.38 & 13.39 & 13.41 & 13.39 \\
    \addlinespace[2pt]
    Qwen2.5-7B & WikiText-2 & 8.63 & 8.60 & 8.62 & 8.62 & 7.89 & 7.40 & 7.41 & 7.43 & 7.42 & 7.41 & 7.40 & 7.37 \\
     & C4 & 15.17 & 15.07 & 15.18 & 15.13 & 14.27 & 13.44 & 13.51 & 13.54 & 13.52 & 13.51 & 13.47 & 13.39 \\
    \addlinespace[2pt]
    Llama-3.1-8B & WikiText-2 & 10.30 & 10.27 & 10.36 & 10.40 & 10.37 & 5.96 & 5.97 & 5.97 & 5.97 & 5.97 & 5.97 & 5.96 \\
     & C4 & 17.30 & 16.66 & 17.37 & 17.38 & 17.54 & 9.13 & 9.13 & 9.13 & 9.13 & 9.13 & 9.13 & 9.12 \\
    \addlinespace[2pt]
    Qwen2.5-14B & WikiText-2 & 5.52 & 5.60 & 5.53 & 5.52 & 5.40 & 5.10 & 5.11 & 5.12 & 5.10 & 5.10 & 5.11 & 5.10 \\
     & C4 & 11.10 & 11.18 & 11.10 & 11.12 & 11.11 & 10.81 & 10.81 & 10.84 & 10.81 & 10.81 & 10.81 & 10.81 \\
    \addlinespace[2pt]
    Qwen3-30B-A3B & WikiText-2 & 8.67 & 8.91 & 8.66 & 8.68 & 8.50 & 8.39 & 8.39 & 8.38 & 8.36 & 8.38 & 8.37 & 8.40 \\
     & C4 & 15.61 & 16.29 & 15.66 & 15.54 & 14.65 & 14.37 & 14.39 & 14.38 & 14.38 & 14.37 & 14.37 & 14.37 \\
    \bottomrule
  \end{tabular}%
  }
  \vspace{3pt}
\end{table}

Table~\ref{tab:full_ablation} shows that design choices have little effect at W16A16: all evaluated settings remain within approximately 1.2\% of the BF16 reference PPL.
At W12A12, however, activation scale selection becomes substantially more important.
Dynamic per-token quantization improves PPL over the static baseline in 17 of the 18 model--dataset pairs.
For example, Llama-3.1-8B's C4 PPL decreases from 17.30 to 9.13, close to the BF16 reference of 9.12.
Changing only the scale representation does not recover this loss, while asymmetric quantization provides model-dependent improvements.
These results suggest that the degradation at W12A12 is not solely determined by bit width: adapting activation scales can use the available precision more effectively.
However, the dynamic setting changes both scale granularity and runtime adaptation, so this experiment does not isolate their individual effects or establish their proving-cost implications.

\section{Use Case: Proving Training-Data (Non-)Membership in Zero-Knowledge}
\label{sec:membership}

In this section, we demonstrate a ZK-LLM governance use case with our ZK-friendly quantization for mandated training-data disclosure (EU AI Act~\citealp{euaiact}, GDPR Article 15~\citealp{gdpr}): proving whether a given text was likely included in a model's training data without revealing the model itself.

In this setting, a model provider can use a ZKP to certify the result of a membership-inference computation, based on the Min-K\% probability score \cite{shi2023detecting}, while keeping the model parameters and intermediate inference values private.

Specifically, given a text $X$, Min-K\% Prob computes $\log
P(x_i \mid x_{<i})$ for every token of $X$, takes the K\% tokens with the
smallest log-probability, and computes their mean. A score above a calibrated
threshold $\varepsilon$ suggests that $X$ was seen during training. 
To prove this score in ZK, the prover needs to prove the LLM inference on $X$ to obtain the probability of each token, followed by the algorithm above. As there are already multiple prior works on LLM inference~\cite{sun2024zkllm,qu2025zkgpt}, we focus on the Min-K\% Prob score part. 

\paragraph{Protocol.} Our protocol is described in Figure~\ref{fig:mink_protocol}. We present the protocol in terms of lookup operations and modular arithmetic, so that the protocol is agnostic to the underlying ZK protocol. The protocol takes the quantized per-token probability vector computed by the ZK inference of the LLM as a secret witness. It computes the fixed-point negative log-probabilities using a precomputed lookup table, validates the sorted vector $\ell^\star$, computes the sum of the $k$ entries of the least likely tokens, and compares the sum with the threshold. Only the membership decision is revealed, while the score stays private. All computations can be realized in ZK protocols efficiently. 

\begin{figure}[htbp]
\centering
\fbox{\begin{minipage}{0.96\textwidth}
\small
\textbf{Public input:} sequence length $N$, Min-K parameter $k$ with $0<k<N$, threshold $T$,
and preprocessed lookup tables $T_{\log}$ and $T_{\mathrm{range}}$. For $q \in
[1,2^{16})$, the log table maps $q$ to the fixed-point negative log-probability
$\lfloor -\log(q/2^{16})\cdot 2^{15}\rceil$, which is below $2^{19}$. The range
table is $T_{\mathrm{range}}=[0,2^{16})$ denoting non-negative numbers.\\
\textbf{Secret Witness:} secret per-token quantized probability vector $q=(q_0,\ldots,q_{N-1})
\in [1,2^{16})^N$ from the quantized model's forward pass on text $X$.\\
\textbf{Auxiliary input:} negative log-probabilities $\ell \in \mathbb{F}^N$, sorted copy $\ell^{\star} \in
\mathbb{F}^N$, table
multiplicities for $T_{\log}$ and $T_{\mathrm{range}}$ used in lookup arguments. The accumulator's final
value $s_{\mathrm{final}}$ stays private inside the committed trace.\\
\textbf{Output:} $\mathsf{accept}/\mathsf{reject}$.
\begin{enumerate}[leftmargin=2em,itemsep=0.2em]
    \item The protocol computes the negative log-probability of each token via lookup operations:
    \[
        \ell_i=\texttt{Lookup}(T_{\log},q_i).
    \]
    \item The protocol checks that $\ell^\star$ and $\ell$ are permutations of each other via a permutation argument. 

    \item The protocol checks that $\ell^\star$ is sorted in descending order. That is, define $d_i = \ell^{\star}_{i}-\ell^{\star}_{i+1}$ for $i=0,\ldots,N-2$ and check $d_i \in [0,2^{24})$ via lookups of its limbs in $T_{\mathrm{range}}$. Since $\ell^\star_i<2^{19}$, an unsorted pair would give $d_i \ge p-2^{19}$ in $\mathbb{F}_p$, outside this range.
    \item The protocol sums the $k$ entries of the least likely tokens: $
        s_{\mathrm{final}}=\sum_{j=0}^{k-1}\ell^{\star}_j.
    $
    \item If $s_{\mathrm{final}}<T$, output \texttt{accept}; otherwise, output \texttt{reject}. With $T=\lceil -k\varepsilon\cdot 2^{15}\rceil$, this is the Min-K\% test of mean log-probability above $\varepsilon$.
\end{enumerate}
\end{minipage}} \caption{Protocol for the Min-K\% membership inference
zero-knowledge proof. }
\vspace{-15pt}
\label{fig:mink_protocol}
\end{figure}

\paragraph{Implementation. }
We implement our protocol using the Plonky3~\citep{plonky3} library. The performance is measured using its
\texttt{p3-batch-stark} prover over the BabyBear field with a Poseidon2-based FRI commitment scheme. As baselines, we implement the same algorithm as two programs for the SP1 zkVM~\citep{sp1zkvm}: one computes the Min-K\% Prob score directly in FP32 without ZK-friendly quantization, and the other uses our fixed-point encoding. The two programs differ only in number representation. In all three implementations, the prover supplies the sorted order, the proof checks it in linear time, and only $(N, k, T)$ and the decision are public. All implementations use the same FRI parameters as the SP1 core prover and are measured on a single socket
of an Intel Xeon Gold 6526Y (16 cores, 256GB DDR5).

\paragraph{Results.}
Table~\ref{tab:mink_zk_numbers} summarizes the proving cost.
Within the same zkVM, ZK-friendly quantization reduces the number of executed RISC-V cycles compared with FP32, because each logarithm becomes a single table lookup instead of a software floating-point routine. 
For short texts, both programs are dominated by the fixed cost of the zkVM (about 25 seconds), but the gap in proving time grows with the text length.
Our protocol proves the decision in about 3 seconds for every $N$ up to 4096.
This is faster than the fixed-point zkVM program and up to 30 times faster than the FP32 program.
The cost is dominated by committing the two $2^{16}$-entry lookup tables, so it barely depends on $N$.
Moreover, as shown in Table~\ref{tab:logit_fidelity_exact}, the W16A16 pipeline preserves near 98\% of the bottom-20\% token positions selected by the BF16 model, which are exactly the tokens that Min-K\% aggregates. 
Hence, the proved decision is computed over nearly the same tokens as the original model would select. 

In conclusion, this use case again confirms that ZK-friendly quantization lowers the cost of proving the Min-K\% statistic. 
It makes the computation expressible as lookups and field arithmetic, allowing our protocol to be proved in a few seconds regardless of the text length.

\begin{table}[t]
  \centering
  \caption{Cost of proving the Min-K\% decision with and without ZK-friendly quantization
  (mean of three proofs).}
  \small
  \setlength{\tabcolsep}{4pt}
  \begin{tabular}{rrrrrr}
    \toprule
    & \multicolumn{1}{c}{\textbf{Ours (Plonky3)}}
    & \multicolumn{2}{c}{\textbf{SP1, fixed point}}
    & \multicolumn{2}{c}{\textbf{SP1, FP32}} \\
    \cmidrule(lr){2-2}\cmidrule(lr){3-4}\cmidrule(lr){5-6}
    $N$ & Prover (s) & Prover (s) & Cycles & Prover (s) & Cycles \\
    \midrule
      64 & 3.15 & 25.61 &  12,794 & 25.29 &   107,677 \\
     128 & 3.59 & 26.00 &  17,863 & 27.11 &   207,150 \\
     256 & 3.28 & 25.97 &  28,001 & 30.50 &   405,742 \\
     512 & 2.98 & 26.48 &  48,268 & 35.58 &   806,221 \\
    1024 & 3.58 & 27.33 &  88,802 & 44.61 & 1,622,618 \\
    2048 & 3.17 & 29.43 & 169,879 & 61.29 & 3,233,201 \\
    4096 & 3.08 & 32.99 & 332,033 & 91.24 & 6,460,814 \\
    \bottomrule
  \end{tabular}
  \label{tab:mink_zk_numbers}
\end{table}

\section{Detailed Experimental Results}
\label{sec:detailed_results}

\begin{table}[p]
  \centering
  \caption{
    Complete operator-wise LUT ablations.
    All values are perplexity ratios relative to the corresponding
    W16A16 LUT-off baseline.
    O: only the indicated operator family uses a B16 LUT;
    all other nonlinear functions are exact.
    E: the full B16 pipeline uses the exact function only for
    the indicated operator family.
  }
  \label{tab:lut-diagnosis-full}
  \scriptsize
  \setlength{\tabcolsep}{2pt}
  \renewcommand{\arraystretch}{1.10}
  \begin{tabular*}{\textwidth}{
    @{\extracolsep{\fill}}lc*{8}{r}@{}
  }
    \toprule
    & & \multicolumn{2}{c}{Llama-3.1-8B}
        & \multicolumn{2}{c}{Qwen2.5-7B}
        & \multicolumn{2}{c}{Qwen2.5-14B}
        & \multicolumn{2}{c}{Qwen3-30B-A3B} \\
    \cmidrule(lr){3-4}
    \cmidrule(lr){5-6}
    \cmidrule(lr){7-8}
    \cmidrule(lr){9-10}
    Operator / configuration & Mode
      & Wiki & C4 & Wiki & C4 & Wiki & C4 & Wiki & C4 \\
    \midrule
    Full B16 & --
      & 1126.8453 & 380.5457
      & 1496.9087 & 769.6439
      & 32.4609 & 18.2997
      & 131.3352 & 120.6984 \\
    \midrule
    RMS InvSqrt & O
      & 1126.7972 & 380.6613
      & 1528.4988 & 759.8001
      & 32.6139 & 18.1795
      & 1.0398 & 1.0562 \\
    & E
      & 0.9993 & 0.9997
      & 0.9999 & 0.9978
      & 1.0002 & 0.9997
      & 1.0034 & 0.9995 \\
    \addlinespace[2pt]
    SiLU & O
      & 0.9999 & 1.0005
      & 0.9998 & 0.9993
      & 0.9998 & 1.0003
      & 1.0038 & 0.9992 \\
    & E
      & 1117.7300 & 382.1309
      & 1415.1683 & 757.1557
      & 31.7131 & 18.5148
      & 132.4172 & 116.0887 \\
    \addlinespace[2pt]
    Attention Exp & O
      & 0.9997 & 0.9996
      & 1.0007 & 1.0010
      & 0.9998 & 1.0006
      & 1.0018 & 0.9984 \\
    & E
      & 1127.2139 & 382.8649
      & 1444.9728 & 755.7837
      & 32.6841 & 18.0263
      & 126.6643 & 112.9952 \\
    \addlinespace[2pt]
    Attention Reciprocal & O
      & 1.0000 & 1.0000
      & 0.9998 & 0.9995
      & 1.0000 & 0.9997
      & 0.9994 & 1.0002 \\
    & E
      & 1123.3660 & 381.7359
      & 1409.3671 & 746.2264
      & 33.1306 & 18.4140
      & 131.2997 & 113.2920 \\
    \midrule
    Q/K RMS InvSqrt & O
      & -- & -- & -- & -- & -- & --
      & 1.0008 & 0.9998 \\
    & E
      & -- & -- & -- & -- & -- & --
      & 1.0399 & 1.0629 \\
    \addlinespace[2pt]
    Router Exp & O
      & -- & -- & -- & -- & -- & --
      & 0.9988 & 0.9989 \\
    & E
      & -- & -- & -- & -- & -- & --
      & 132.8450 & 115.8493 \\
    \addlinespace[2pt]
    Router Reciprocal & O
      & -- & -- & -- & -- & -- & --
      & 0.9989 & 1.0000 \\
    & E
      & -- & -- & -- & -- & -- & --
      & 127.8957 & 123.5119 \\
    \addlinespace[2pt]
    Router Top-$k$ Reciprocal & O
      & -- & -- & -- & -- & -- & --
      & 0.9995 & 1.0011 \\
    & E
      & -- & -- & -- & -- & -- & --
      & 128.5913 & 126.5326 \\
    \midrule
    Full B16; RMS B24 & --
      & \textbf{1.0002} & \textbf{1.0025}
      & \textbf{0.9987} & \textbf{0.9906}
      & \textbf{0.9997} & \textbf{1.0006}
      & \textbf{1.0014} & \textbf{0.9994} \\
    \bottomrule
  \end{tabular*}
\end{table}

\begin{table}[t]
  \centering
  \caption{LUT cost sensitivity on GPT-2 Small.}
  \label{tab:lut-cost}
  \small
  \setlength{\tabcolsep}{3pt}
  \renewcommand{\arraystretch}{1.12}
  \begin{tabular*}{\textwidth}{
    @{\extracolsep{\fill}}crrrrrrr@{}
  }
    \toprule
    & \multicolumn{3}{c}{Proving time (seconds)}
    & \multicolumn{4}{c}{Analysis: GeLU $B$, Exp $B$} \\
    \cmidrule(lr){2-4}
    \cmidrule(lr){5-8}
    $B$
      & \shortstack{GeLU $B$\\Exp 16}
      & \shortstack{GeLU 16\\Exp $B$}
      & \shortstack{GeLU $B$\\Exp $B$}
      & \shortstack{LUT payload\\(MiB)}
      & \shortstack{Direct queries\\(M)}
      & \shortstack{All queries\\(M)}
      & \shortstack{Joint table\\proof (s)} \\
    \midrule
    8
      & 103.58 (0.77)
      & 104.70 (0.53)
      & 104.75 (0.37)
      & 0.102 & 0.823 & 18.238 & 0.483 \\
    12
      & 103.37 (0.34)
      & 104.47 (1.25)
      & 103.28 (0.73)
      & 1.625 & 0.823 & 17.906 & 0.592 \\
    16
      & 102.65 (3.04)
      & 102.65 (3.04)
      & 102.65 (3.04)
      & 26.000 & 0.823 & 17.104 & 1.325 \\
    24
      & 248.44 (3.51)
      & 120.49 (1.13)
      & 265.54 (2.93)
      & 6,656.000 & 0.823 & 18.218 & 103.761 \\
    \bottomrule
  \end{tabular*}

  \vspace{3pt}
  \begin{minipage}{\textwidth}
    \footnotesize
    Times are means (standard deviation) over three runs.
    At $B=16$, all three columns use the same configuration.
  \end{minipage}
\end{table}

\begin{table}[t]
  \centering
  \caption{Linear proving costs for one Qwen2.5-14B block.}
  \label{tab:qwen-linear-proving}
  \small
  \setlength{\tabcolsep}{3pt}
  \renewcommand{\arraystretch}{1.08}
  \begin{tabular*}{\textwidth}{
    @{\extracolsep{\fill}}llrrrr@{}
  }
    \toprule
    & & \multicolumn{2}{c}{Proving time (seconds)}
      & \multicolumn{2}{c}{Peak RAM usage (GiB)} \\
    \cmidrule(lr){3-4}
    \cmidrule(lr){5-6}
    Projection
      & Weight shape
      & Floating-point
      & Integer
      & Floating-point
      & Integer \\
    \midrule
    \multicolumn{6}{@{}l}{
      \textit{Direct measurements}
    } \\
    Q/K/V/O, gate/up
      & $5120 \times 8$
      & 18.998 & 1.115
      & 14.8816 & 0.2048 \\
    FFN down
      & $13824 \times 8$
      & 42.352 & 1.359
      & 40.7634 & 0.2943 \\
    \midrule
    \multicolumn{6}{@{}l}{
      \textit{Full-width costs: floating-point estimated, integer measured}
    } \\
    Q
      & $5120 \times 5120$
      & 12,158.78 & 11.877
      & -- & -- \\
    O
      & $5120 \times 5120$
      & 12,158.78 & 11.877
      & -- & -- \\
    K
      & $5120 \times 1024$
      & 2,431.76 &  4.603
      & -- & -- \\
    V
      & $5120 \times 1024$
      & 2,431.76 &  4.603
      & -- & -- \\
    FFN gate
      & $5120 \times 13824$
      & 32,828.71 & 20.985
      & -- & -- \\
    FFN up
      & $5120 \times 13824$
      & 32,828.71 & 20.985
      & -- & -- \\
    FFN down
      & $13824 \times 5120$
      & 27,105.54 & 20.526
      & -- & -- \\
    \midrule
    \multicolumn{2}{l}{\textbf{One-block linear total}}
      & \textbf{121,944.05}
      & \textbf{95.457}
      & -- & -- \\
    \bottomrule
  \end{tabular*}
  \vspace{3pt}
\end{table}

\end{document}